# An associative memory model with very high memory rate: Image storage by sequential addition learning

Hiroshi Inazawa

*Center for Education in Information Systems, Kobe Shoin Women's University,*
*1-2-1 Shinohara-Obanoyama, Nada Kobe 657-0015, Japan.*

## Abstract

In this paper, we present a neural network system related to about memory and recall that consists of one neuron group (the "cue ball") and a one-layer neural net (the "recall net"). This system realizes the bidirectional memorization learning between one cue neuron in the cue ball and the neurons in the recall net. It can memorize many patterns and recall these patterns or those that are similar at any time. Furthermore, the patterns are recalled at most the same time. This model's recall situation seems to resemble human recall of a variety of similar things almost simultaneously when one thing is recalled. It is also possible for additional learning to occur in the system without affecting the patterns memorized in advance. Moreover, the memory rate (the number of memorized patterns / the total number of neurons) is close to 100%; this system's rate is 0.987. Finally, pattern data constraints become an important aspect of this system.

## 1. Introduction

We present a model that realizes bidirectional memorization learning between one neuron group i.e., the "cue ball," and one neural net i.e., the "recall net," which shows that the system recall memorized patterns or similar patterns at any time. In 2018, the mechanism of memorization learning for the recall net was reported in detail (Inazawa,2018). This previous report, showed that additional learning is possible without affecting previous memorized patterns, and that memory rate (the number of memorized patterns / the number of total neurons in a system) approaches 100% (specifically 0.987) as the number of neurons in the system increases. In the present study, we focus on how the system can learn bidirectionally between the cue ball and recall net and how the memorized patterns can be recalled by stimulating external patterns on neurons in the cue ball.

In a simulation, we show that the output of one learned neuron in the cue ball becomes larger than the outputs of other neurons when a memorized pattern is presented on the recall net and that the large output of the cue neuron leads to recall of the memorized pattern on the recall net. In this series of processes, there are many patterns similar to the presented memorized pattern. This situation seems to resemble human recall in which a variety of similar things is recalled at about the same time when one thing is recalled.

Memory models in neural networks are an important field of study. We assessed associative memory models were during the second neural network boom, and many studies of associative memory have been reported so far (Nakano, 1972, Anderson, 1972, Kohonen, 1972, Amari, et al., 1988, Yoshizawa et al., 1993, Inazawa et al., 1998). At the time of those studies storage capacity (memory rate) was a major problem in these models; thus, improving this capacity was of the research. Many important studies have shown that memory rate is approximately 15%~50% of the number of neurons in an output layer (Hopfield, 1982, Amit et al., 1985, Amari et al., 1988, Yoshizawa et al., 1993). Another problem in these models was the act of additional learning processes destroying previously learned data. In a previous report (Inazawa, 2018), we showed how these problems could be improved: our new model achieved additional learning without destroying previous data, and greatly improved memory rate.

In Section 2, we describe the specifications of the current model. In section 3, we present the method of simulation and its results. In Section 4, we summarize our findings and discuss potential future research.

2. Specification of the model

The proposed model consists of two parts, i.e., cue ball and recall net. The cue ball is a group of many cue neurons, whereas the recall net is a one-layer neural net consisting of a fixed number of recall neurons, which is shown in Figure 1. Notably, the cue ball can be any shape.

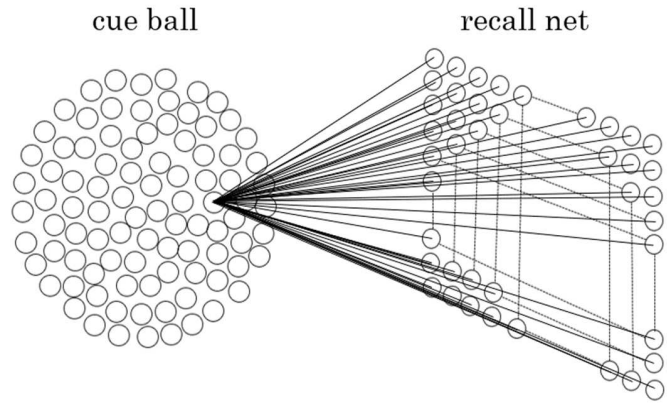

Figure 1: Proposed neural network system. The cue ball consists of many cue neurons while the recall net consists of a fixed number of recall neurons. One cue neuron is connected to all recall neurons with bidirectional learning occurring between these neurons.

For example, although the cue ball is represented as spherical here, it can also be represented as a chain structure, which would be created by adding one cue neuron each time the pattern memorization increases. In the previous report (Inazawa, 2018), we named the cue neuron the "trigger neuron." In the current study, we renamed this neuron the cue neuron. Memorization learning is executed bidirectionally between one cue neuron[1] and many recall neurons.

Although the cue neuron is connected to all recall neurons, it shares only a threshold, but no internal connections with the cue ball's cue neurons. Similarly, recall neurons are also connected to all cue neurons, but they do not share a threshold and no internal connections between recall neurons. Although the number of cue neurons can increase infinitely, we prepared 60,000 cue neurons according to the number of patterns used in the simulation. On the other hand, the number of recall neurons is 784 according to the number of pixels of a pattern (28 × 28 pixels). When processing occurs from a cue neuron to the recall neurons, the cue neuron has one input and multiple outputs, whereas a recall neuron has multiple inputs and one output. On the other hand, when the process is reversed, the cue neuron has multiple inputs and one output, whereas a recall neuron has one input and multiple outputs.

The working processes of the system are follows. The processes consist of two parts: the memorization learning process and recall process. When these processes are complete, the recall net reminded of the memorized patterns using learned cue neurons' output.

The learning process is as follows. First, the recall neurons learn based on the connection weights to the recall neurons and then, it for the cue neurons learn based on the connection weights to the cue neuron. The gradient descent method (GDM) was used as the learning method (Widrow et al., 1960, Rescorla et al., 1972, Rumelhart et al., 1986). We begin with the learning of the recall neurons. One cue neuron from the cue ball is picked it up. For the recall neurons' memorization learning, the pattern data is presented to the recall net as the teacher signal. Note that the data are converted from pixel values into grayscale data. After all recall neurons have learned, the outputs of all recall neurons can be obtained by using the output of the previously selected cue neuron. The output values of the recall neurons are then returned to the selected cue neuron. Subsequently, the memorization learning is executed for the cue neuron using the preset cue neuron's teacher signal. Bidirectional learning between the cue neuron and recall neurons is completed via these processes.

---

[1] A grandmother cell has previously been proposed, i.e., where one neuron is responsible for memory.

The recall process is as follows. An appropriate memorized pattern is presented on the recall net, and the data of this pattern enter in the cue neurons. When the cue neuron's output becomes the teacher signal or a signal close to that value, it can be said that the cue neuron has learned the presented pattern. Afterward, when the cue neuron's output (normalized to 1.0) enter all recall neurons, the memorized pattern is visualized on the recall net. On the other hand, multiple cue neurons are nominated as neurons that have learned the presentation pattern by adjusting the cue neuron's threshold. In this case, we may also see multiple patterns similar to the memorized pattern.

Next, we describe the model in detail. First, we explain the input-output process from the cue neuron to the recall neurons and the learning process for the recall neurons' connection weights. The input-output relation is as follows:

$$y_j^p = w_{ji}^p x_i^p , \qquad (1)$$

where $w_{ji}^p$ is the connection weight from *i*th cue neuron to *j*th recall neuron, *i* is the cue neuron number, and *j* is the recall neuron number. Note that in equation (1), there is no the summation of *i*. $x_i^p$ denotes the cue neuron's output of the *i* th, where $i = 0 \sim 59{,}999$, and $p = 0 \sim 59{,}999$ indicates the pattern number, respectively: the values of *i* and *p* are equal in the simulation. $y_j^p$ denotes the output of the *j*th recall neuron, where $j = 0 \sim 783$. We define the error function of the recall neurons as follows:

$$E^p \equiv \frac{1}{2}\sum_{j=0}^{M}\left(d_j^p - y_j^p\right)^2 , \tag{2}$$

where $d_j^p$ is the element value (grayscale data) of the pattern and becomes the teacher signal for $w_{ji}^p$. $M$ is the last number of recall neurons, i.e., 783. The update equations of $w_{ji}^p$ updated by GDM are as follows:

$$w_{ji}^p(t+1) = w_{ji}^p(t) + \Delta w_{ji}^p(t) , \tag{3}$$

$$\Delta w_{ji}^p(t) \equiv -\varepsilon_W \frac{\partial E^p}{\partial w_{ji}^p(t)} = \varepsilon_W \left(d_j^p - y_j^p(t)\right) x_i^p(t) , \tag{4}$$

where $t$ denotes the update count, and $\varepsilon_W$ is the learning rate. Using the input-output equation (1) and the learning equations (3) and (4), the recall neuron can certainly learn the teacher signal of the pattern. Let's actually calculate the equation (1) with $t+1$ and set $\varepsilon_W$ to 1.

$$\begin{aligned}y_j^p(t+1) &= w_{ji}^p(t+1)x_i^p(t+1)\\ &= (w_{ji}^p(t) + \left(d_j^p - w_{ji}^p x_i^p(t)\right)x_i^p(t))x_i^p(t+1).\end{aligned} \tag{5}$$

Considering the output of $i$ th cue neuron is set as $x_i^p(t) = x_i^p(t+1) = 1$, equation of (5) can be transformed as follows:

$$y_j^p(t+1) = d_j^p . \tag{6}$$

After the learning, the output of $y_j^p$ is equal to the teacher signal. Next, we explain the input-output process from the recall neuron to the cue neuron and the learning process for the connection weights of the cue neuron. The input-output relation is as follows:

$$x_i^p = f(q_i^p) = f\left(\sum_{j=0}^{M} v_{ij}^p y_j^p\right) = \begin{cases} 0 & for\ q_i^p < H \\ 1 & for\ q_i^p \geq H \end{cases}, \tag{7}$$

where $v_{ij}^p$ is the connection weight from *j*th recall neuron to *i*th cue neuron, where *i* and *j* denote the cue neuron and the recall neuron number, respectively. Here *p* indicates the pattern number, and *H* denotes the threshold value of a cue neuron. $y_j^p$ and $x_i^p$ denote the output of the *j*th recall neuron and the *i*th cue neuron, respectively. The learning uses the same method as that used for the recall neurons. We define the error function of the cue neurons as follows:

$$e^p \equiv \frac{1}{2}(\theta - q_i^p)^2, \tag{8}$$

where $\theta$ is the teacher signal for the cue neurons. The update equations of $v_{ij}^p$ updated by GDM are as follows:

$$v_{ij}^p(t'+1) = v_{ij}^p(t') + \Delta v_{ij}^p(t'), \tag{9}$$

$$\Delta v_{ij}^p(t') = -\varepsilon_V \frac{\partial e^p}{\partial v_{ij}^p} = \varepsilon_V \left(\theta - q_i^p(t')\right) y_j^p(t'), \tag{10}$$

where *t'* denotes the update count, and $\varepsilon_V$ is the learning rate. Using the input-output equation (7) and the learning equations (9) and (10), the cue neuron can certainly learn the teacher signal ($\theta$) of the pattern. Let's actually calculate equation (7) with $t'+1$ and set $\varepsilon_V$ to 1. Also, $y_j^p(t')$ is obtained from equation (6) after the learning of $w_{ji}^p$, $y_j^p = d_j^p$. The result are as follows.

$$q_i^p(t'+1) = \sum_{j=0}^{M} v_{ij}^p(t'+1)y_j^p$$

$$= \sum_{j=0}^{M} v_{ij}^p(t')y_j^p \left(1 - \sum_{j=0}^{M}(y_j^p)^2\right) + \theta \sum_{j=0}^{M}(y_j^p)^2 \quad (11)$$

Here, we set the following restriction on the element values (teacher signals) of the pattern.

$$\sum_{j=0}^{M}(d_j^p)^2 \equiv 1, \quad (12)$$

where equation (12) also be expressed from equation (6) as follows:

$$\sum_{j=0}^{M}(y_j^p)^2 \equiv 1. \quad (13)$$

As the result, equation (11) becomes as follows:

$$q_i^p(t'+1) = \theta \ . \quad (14)$$

Thus, $q_i^p(t'+1)$ is equal to the teacher signal.

3. Simulation and results

Here, we execute the learning and recall processes introduced above via a simulation. In the simulation, we use the data set "train-images-idx3-ubyte" from the MNIST[2] database, consisting of 60,000 images of handwritten characters. Since the patterns are black and white images, we convert these to 8-bit grayscale data. Furthermore, the converted pattern data are normalized according to equation (12), where these data are denoted as $d_j^p$. The initial output values of the cue neurons and recall neurons set to 0.0, and the learning coefficients $\varepsilon_W$ and $\varepsilon_V$ are set to 1.0, respectively.

First, we execute the learning processes with 60,000 cue neurons prepared in the cue ball and 784 recall neurons in the recall net. The patterns to be used are picked up one by one from *0* th to *m-1*th of the data set, where *m* is 1,000. After learning up to the *m-1*th, the pattern is picked up one by one from the *m* th to *2m-1*th, and then the learning is performed again. The learning algorithm is as follows:

1. Initialize the connection weights of the recall neurons: $w_{ji}^p = 1.0$
2. Picks one (*i*th) cue neuron up from the cue ball and set the output value of this cue neuron: $x_i^p = 1.0$
3. Calculate the $y_j^p$ using $w_{ji}^p$ and $x_i^p$
4. Learn $w_{ji}^p$ using teacher signal $d_j^p$
5. Calculate $y_j^p$ again for the *i*th cue neuron using learned $w_{ji}^p$
6. Initialize the connection weights of the cue neurons: $v_{ij}^p = 1.0$
7. Calculate $q_j^p$ for *i* th cue neuron using $v_{ij}^p$ and $y_j^p$
8. Learn $v_{ij}^p$ using the teacher signal $\theta$, where $\theta$ is set to 100.

---

[2] http://yann.lecun.com/exdb/mnist/.

In addition, $E^p$ and $e^p$ become 0 once learning.

In the recall processes, after the memorized pattern data has been inputted to the learned cue neurons, we identify a cue neuron for which $q_i^p$ is equal to or close to the teacher signal $\theta$(=100). If this cue neuron is found, we set $x_i^p$ to 1.0 using the threshold $H$ of the cue neurons and enter it for all recall neurons. The algorithm of the recall process is as follows:

1. Select the memorized pattern
2. Present the pattern data $d_j^p$ to the recall net
3. Enter the output values to all cue neurons and calculate $q_i^p$
4. When $q_i^p \geq H(= 90)$, the output of this (ith) cue neuron sets to 1.0: $x_i^p = 1.0$
5. Calculate the outputs of recall neurons using the output of this cue neuron

In this process, we observe patterns similar to the presented pattern by adjusting the cue neuron threshold $H$.

We also examined the difference in shape and shading between the recall pattern and the memorized pattern. We determined the hamming distance between these patterns by converting each pixel value to a digital value for the difference in shape. Results showed that the hamming distance was 0 for 60,000 patterns. Thus, we conclude that the shapes between these patterns are identical. On the other hand, we determined the distance between each pixel value (grayscale data) of the patterns for the difference in shading. The average distance was 2.19 for 60,000 patterns, where the average total pixel value (grayscale value) of one pattern for 60,000 patterns was 26,210.05. Therefore, we conclude that the shades between the patterns are almost completely recalled.

4. Conclusions and discussion

The system proposed in this paper consists of one neuron group known as a cue ball and a one-layer neural network known as a recall net. The presented model can learn bidirectionally between one cue neuron and the recall neurons, memorize many patterns, and recall those patterns or similar pattern. After learning, when learned patterns are presented to the cue neurons, the corresponding cue neuron that has learned and memorized the pattern outputs the teacher signal or a value close to the signal. The memorized pattern is recalled by using the output value of this cue neuron at the recall net. We also found that the hamming distance between the presented patterns and recall patterns was all of 0 for 60,000 patterns. On the other hand, we found small difference in the shading distance between the presented patterns and recall patterns, i.e., the average distance for 60,000 patterns was 2.19. Therefore, we conclude that the shape and shade between the patterns are almost completely recalled. In addition, the memory rate has been 0.987 in this model.

An important aspect of the cue neuron's recall process is the normalization of presented pattern data, as shown in equation (12). The cue neuron that memorizes a pattern is identified through this normalization. This cue outputs the teacher signal ($\theta$=100) or a signal similar to the teacher signal when a memorized pattern is presented. In this case, many outputs smaller than that of the teacher signal appear; however, the closer the signal is to the teacher signal, the more similar the recall pattern is to the presentation pattern as shown in Figures 2 and Figure 3. In Figure 2, we show the firing spectrum of the cue neuron's outputs, where the cue neurons picked up are from 0th to 999th.

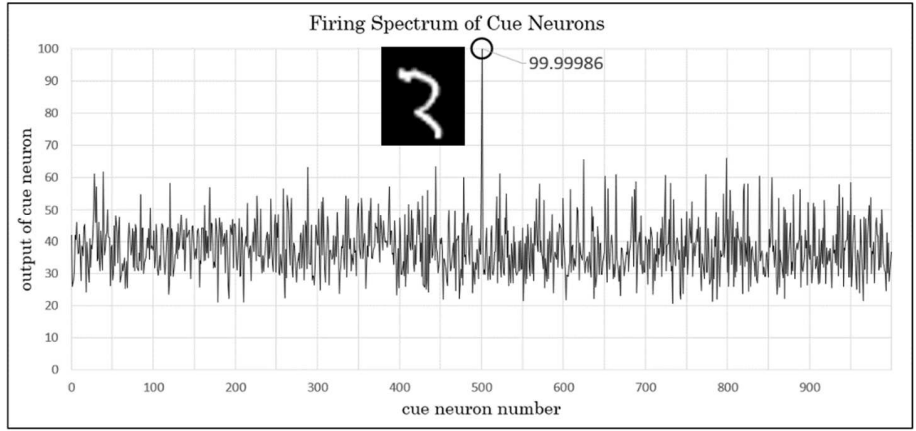

Figure 2: The firing spectrum of cue neurons. Shown are the cue neurons' outputs, with the cue neurons running from 0th to 999th. The vertical axis denotes $q_i^p$. In this case, the memorized pattern of the 500th cue neuron is presented. The peak value indicates 500th cue neuron's output, which is 99.99986. Also, inset image in black rectangle is that of the presented pattern.

The peak value in Figure 2 shows the output of the 500th (counting from 0) cue neuron, for which the presenting pattern number and cue neuron number are the same. The peak value in Figure 2 shows the cue neuron output when presenting the memorized pattern of the 500th cue neuron: the output value is 99.99986 at the 500th cue neuron, which is almost 100 (i.e., the teacher value). Thus, we conclude that the cue neuron recalls the teacher signal almost perfectly. On the other hand, the other cue neuron's output values are much lower than that of the 500th cue neuron. Since the 500th pattern has an unusual shape, the outputs other than the 500th did not grow as similar patterns. Figure 3 presents the memorized patterns of the 600 th cue neuron.

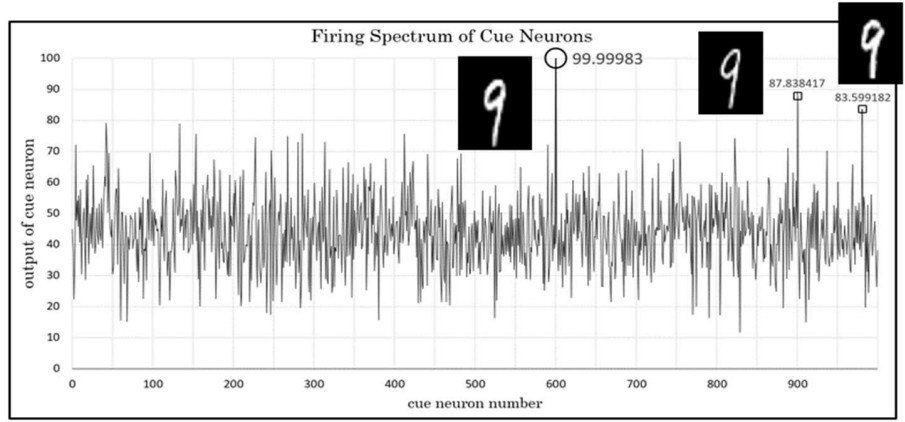

Figure 3: The firing spectrum of cue neurons. Shown are the cue neurons' outputs, with the cue neurons running from 0 th to 999th, and the vertical axis denotes $q_i^p$. In this case, the memorized pattern of the 600th cue neuron is presented. The peak value indicates the 600th cue neuron output, which is 99.99983. The rectangle images inset in the figure show the images the presented patterns, where the output values of these patterns are lower than 80.

The 600th output of 99.99983 is almost equal to the teacher signal ($\theta$=100). If the threshold value *H* is set to 90, the recall pattern contains only this memorized pattern. As shown in Figure 3, when the *H* is set to 80, three recall are obtained (from 600th, 900th, 980th cue neurons): the 900th and 980th patterns have a very similar shape to that of the image for the 600th pattern. The output values of the 900th and 980th patterns were 87.838417 and 83.599182, respectively. It seems that there is no problem with any of these three patterns as the correct recall pattern. On the other hand, if we set the value of *H* lower than 80, so many patterns become recall pattern candidates so that the system lose control. Next, we show a case in which the memorized pattern is partially presented. Figure 4 shows the outputs of cue neurons when the upper half of 500th memorized pattern is presented. The output value is smaller overall, but the 500th peak value is still the largest at 53.1339. Thus, we can conclude that recall can still be achieved by partially presenting the pattern.

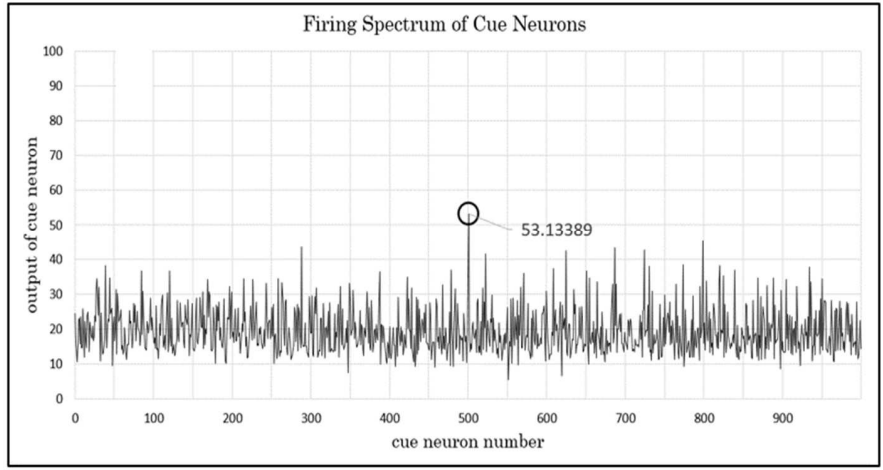

Figure 4: The firing spectrum of cue neurons. Shown are the cue neurons' outputs, with the cue neurons running from 0th to 999th, and the vertical axis denotes $q_i^p$. In this case, the upper half of 500th memorized pattern is presented. The peak value indicates the 500th cue neuron output, which is 53.13389.

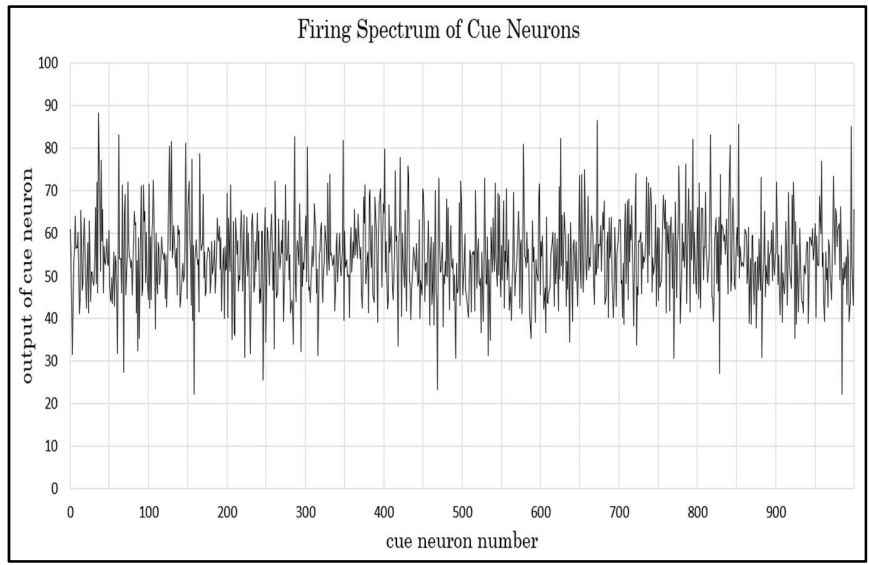

Figure 5a: The firing spectrum of cue neurons. Shown are the cue neurons' outputs, with the cue neurons running from 0th to 999th. The vertical axis shows $q_i^p$. In this case, an unlearned pattern (9500th) is presented.

We show what occurs when an unmemorized pattern is presented. Figure 5a shows the case where such an unmemorized pattern (9500th) is presented. When $H > 90$, recall patterns do not arise; however, when $H > 80$, 16 recall patterns become candidates. The images of the 16 recall patterns when $H > 80$ are shown in Figure 5b. In this figure, the $q_i^p$ of the recall patterns decreases as we move from the right and move down. For example, the $q_i^p$ at the 36th pattern is 88.20, whereas the $q_i^p$ at the 302th pattern is 80.19. The image of the presented pattern (9500th) is "6" and the images of most of the recall patterns are "6". However, the images at the 625th pattern ($q_i^p$=82.16) and the 348 th ($q_i^p$=81.81) are differ from the presented pattern image. Therefore, the unmemorized pattern can recall some similar patterns, but it also recalls different patterns. To select such the recall patterns, we adjust the value of the threshold $H$.

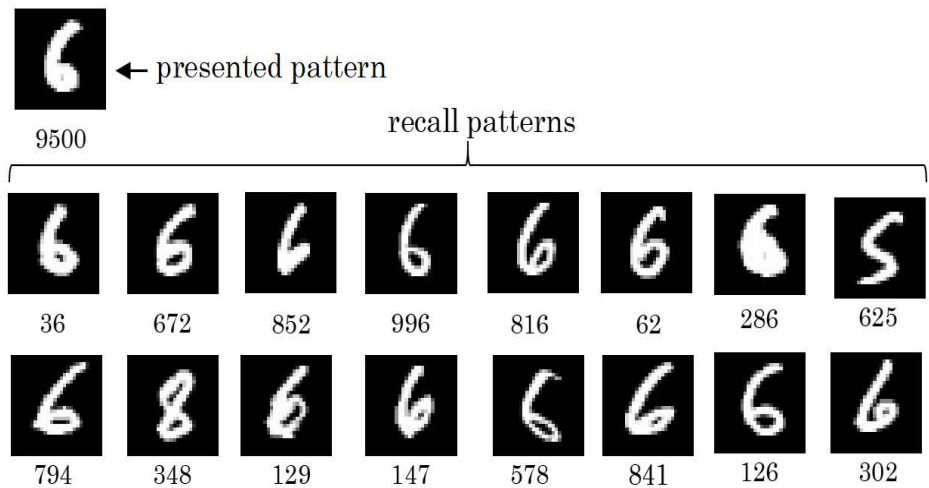

Figure 5b: The images of the recall patterns when $H>80$ and an unmemorized pattern is presented. The numbers under the images are the pattern numbers. The image in the top left is the presented pattern and the others images are the recall patterns.

Finally, we discuss the future development of the proposed model. First, we consider processing video with this model. Since video comprises a series of many images, it is sufficient that the continuous images are displayed on the recall net. To perform this task, the multiple cue neurons and recall neurons need only learn the different consecutive images. After learning, if these cue neurons are fired in sequence, a continuous image will be displayed on the recall net. If the time interval between consecutive images is long, the video will be a time-lapse-like video. Currently, the proposed model has only one recall net; however, but suppose the model has multiple recall nets, as shown in Figure 6.

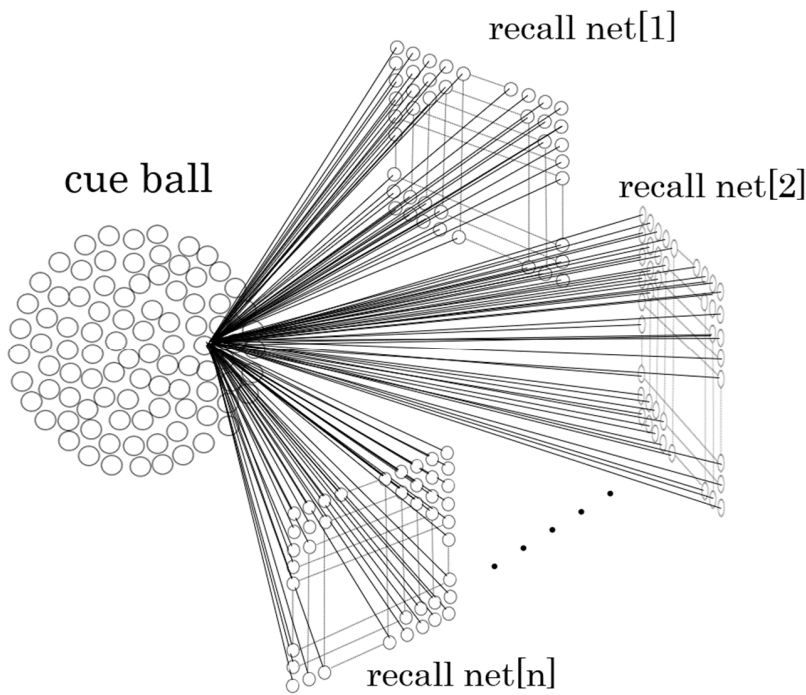

Figure 6: A distributed associative memory model. Recall nets of n-sheets are shown. All recall neurons in the multiple recall nets are connected to one cue neuron in the cue ball.

As shown in Figure 6, all recall neurons from multiple recall nets are connected to one cue neuron in the cue ball. The various patterns presented on these recall nets are memorized to this one cue neuron. After learning has been completed for all recall neurons and the cue neuron, the presentation of a certain pattern in one recall net results in various memorized patterns being displayed on other recall nets connected to this cue neuron. This can be considered a type of associative memory because the various related patterns are recalled from one pattern. Also, suppose that another cue neuron (for example, where two cue neurons are used) is connected to the multiple recall nets. The firing of the first cue neuron causes various patterns to be recalled in several recall nets, which in turn causes the second cue neuron to fire. The firing of the second cue neuron then causes various patterns to appear in the recall net to which it is connected. Consequently, different related recall patterns can be seen in other recall net groups due to firing a second cue neuron. This process is expected to occur on a larger scale if more cue neurons are connected to many recall nets. Hence, this can be understood as a chain reaction of associative memory.

The proposed model and future model discussed here can be considered to have some similarities to human memory and recall. For example, various other memories arise when recalling one memory, while memories that are not directly related are recalled via the chain reaction. Also, the threshold $H$ used in this model may correspond to a means of determining whether the recall is conscious or unconscious. If $H$ varies from person to person, this reflects a person's ability regarding the recall. Moreover, where learned cue neurons fire randomly without presentation of patterns, memorized patterns randomly appear in the recall net. This is a similar situation to the random recall of memories by humans even without information being received into the brain.

In our previous work (Inazawa, 2018), we stated the following: "if the stored data can be recalled in response to external data, it would be interesting in the sense that it resembles the recall of memories we are undergoing." In the current research, we believe that one of the mechanisms has been revealed. As a future research project, it would be interesting to construct an associative memory model that can recall memorized information in a chain reaction.